\documentclass[letterpaper]{article} % DO NOT CHANGE THIS
\usepackage[submission]{aaai23}  % DO NOT CHANGE THIS
\usepackage{times}  % DO NOT CHANGE THIS
\usepackage{helvet}  % DO NOT CHANGE THIS
\usepackage{courier}  % DO NOT CHANGE THIS
\usepackage[hyphens]{url}  % DO NOT CHANGE THIS
\usepackage{graphicx} % DO NOT CHANGE THIS
\usepackage{natbib}  % DO NOT CHANGE THIS AND DO NOT ADD ANY OPTIONS TO IT
\usepackage{caption} % DO NOT CHANGE THIS AND DO NOT ADD ANY OPTIONS TO IT
\usepackage{algorithm}
\usepackage{algorithmic}

\usepackage{newfloat}
\usepackage{listings}
\DeclareCaptionStyle{ruled}{labelfont=normalfont,labelsep=colon,strut=off} % DO NOT CHANGE THIS
\floatstyle{ruled}
\newfloat{listing}{tb}{lst}{}
\floatname{listing}{Listing}
\usepackage{verbatim}
\usepackage{multirow}
\usepackage{booktabs}
\usepackage[table,xcdraw]{xcolor}
\usepackage{amssymb}% http://ctan.org/pkg/amssymb
\usepackage{pifont}% http://ctan.org/pkg/pifont
\usepackage{tikz-cd}
\usepackage{epigraph}
\usepackage{enumitem} 
\usepackage{amsmath}

\title{Learning Action-Effect Dynamics from Pairs of Scene-graphs}
\author{
    Written by AAAI Press Staff\textsuperscript{\rm 1}\thanks{With help from the AAAI Publications Committee.}\\
    AAAI Style Contributions by Pater Patel Schneider,
    Sunil Issar,\\
    J. Scott Penberthy,
    George Ferguson,
    Hans Guesgen,
    Francisco Cruz\equalcontrib,
    Marc Pujol-Gonzalez\equalcontrib
}
\affiliations{
    \textsuperscript{\rm 1}Association for the Advancement of Artificial Intelligence\\
    1900 Embarcadero Road, Suite 101\\
    Palo Alto, California 94303-3310 USA\\
    publications23@aaai.org
}

\usepackage{bibentry}
\begin{document}

\maketitle

\begin{abstract}
`Actions' play a vital role in how humans interact with the world. Thus, autonomous agents that would assist us in everyday tasks also require the capability to perform `Reasoning about Actions \& Change' (RAC). Recently, there has been growing interest in the study of RAC with visual and linguistic inputs. Graphs are often used to represent semantic structure of the visual content (i.e. objects, their attributes and relationships among objects), commonly referred to as scene-graphs. In this work, we propose a novel method that leverages scene-graph representation of images to reason about the effects of actions described in natural language. We experiment with existing CLEVR\_HYP \cite{sampat2021clevr_hyp} dataset and show that our proposed approach is effective in terms of performance, data efficiency, and generalization capability compared to existing models.
%We implement an encoder-decoder architecture to learn the representation of actions as vectors. We combine the aforementioned encoder-decoder architecture with existing modality parsers and a  graph question answering model to evaluate our proposed system on the CLEVR\_HYP dataset. We conduct thorough experiments to 
\end{abstract}

\section{Introduction}

Reasoning about `Actions' is important for humans as it helps us to predict if a sequence of actions will lead us to achieve a desired goal; to explain observations i.e. what actions may have taken place; and to diagnose faults i.e. identifying actions that may have resulted in undesirable situation \cite{baral2010reasoning}. As we are developing autonomous agents that can assist us in performing everyday tasks, they would also require to interact with complex environments. As pointed out by \citet{davis2015commonsense}, imagine a guest asks a robot for a glass of wine; if the robot sees that the glass is broken or has a dead cockroach inside, it should not simply pour the wine. Similarly, if a cat runs in front of a house-cleaning robot, the robot should neither run it over nor put it away on a shelf. Hence, the ability of artificial agents to perform reasoning about actions is highly desirable. 

\begin{figure}[ht!]
\centering
  \includegraphics[width=\linewidth]{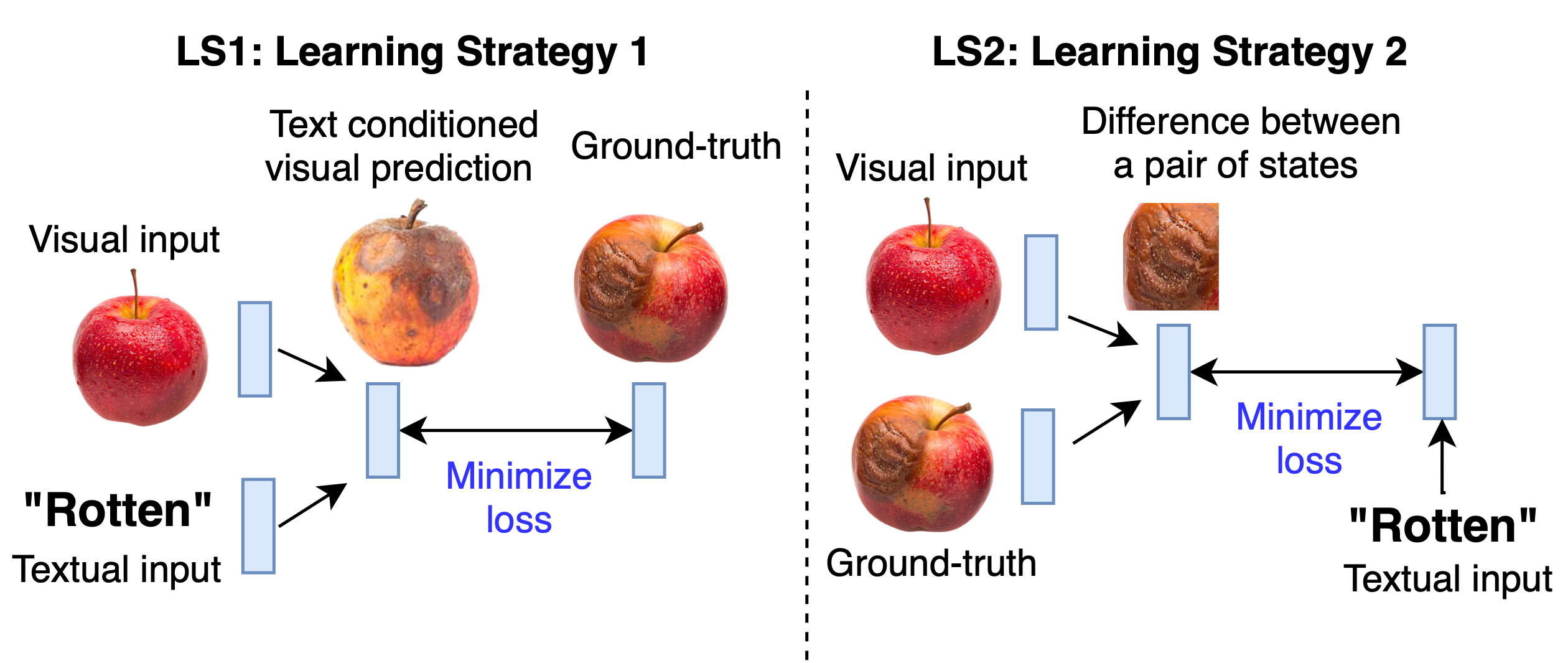}
  \captionof{figure}{{Two possible ways (LS1 and LS2) to learn action-effect dynamics in a supervised learning setting. In this paper, we implement LS2 which demonstrates improvements over LS1. Blue box denotes vector representation.}}
  \label{fig:intuition}
\end{figure}

%As mentioned above, rapid physical inferences are central to how humans interact with the world, and it is important to understand its computational aspects as we make progress in Artificial Intelligence (AI). 
As a result, Reasoning about Action and Change (RAC) has been a long-established research problem, since the rise of Artificial Intelligence (AI). \citet{mccarthy1960programs} were the first to emphasize on reasoning about effects of actions. They developed an advice taker system that can do deductive reasoning about scenarios such as ``going to the airport from home" requires ``walking to the car" and ``driving the car to airport". Since then, many real-life use cases have been identified which require AI models to understand interactions among the states of the world, actions being performed, and most likely following states \cite{banerjee2020can}. 

While RAC has been more popular among knowledge representation and logic community, it has recently piqued the interest of NLP and vision researchers. A recent survey by \citet{sampat2022reasoning} compiled a comprehensive list of  works that explore neural network's ability to reason about actions and changes, provided linguistic and/or visual inputs. Specifically, the works of \citet{park2020visualcomet, sampat2021clevr_hyp, shridhar2020alfred, yang2021visual, gao2018action, patel2022benchmarking} are quite relevant.

In Figure \ref{fig:intuition}, we describe two possible action-effect learning strategies (LS1 and LS2) through a toy example to convey our intuition behind this work. LS1 uses visual features (i.e. features from the image of an apple) and action representation (of text ``rotten'') learned through sentence embedding to imagine effects (i.e. how a rotten apple would look like). LS1 has been an intuitive choice to model action-effect learning in a supervised setting in previous literature. In our hypothesis, LS1 does not improve the model's understanding of what effects the actions will produce. Thus, we propose alternative strategy LS2. Specifically, we let the model observe the difference between pairs of states before and after the action is performed (i.e. decayed portion of the apple that distinguishes a good apple from the rotten one), then associate those visual differences with the corresponding linguistic action descriptions (i.e. text ``rotten''). LS2 is likely to better capture action-effect dynamics, as action representations are learned explicitly.

%is not much helpful as it
%\footnote{Figure \ref{fig:intuition} is meant to convey our intuition behind the proposed model in this work at a very high level. Figure \ref{fig:reasoner} is more complex in comparison and accurately describes the working of our model, considering the format of CLEVR\_HYP dataset, decomposing the task into various neural components, and measuring them using appropriate loss functions.}.

\section{CLEVR\_HYP Dataset \cite{sampat2021clevr_hyp}}
\label{sec:clevr_hyp}  
In this section, we summarize important aspects of CLEVR\_HYP  and terminology used in subsequent sections. 

\subsection{Problem Formulation} 
\begin{figure}[ht!]
 \setlength\belowcaptionskip{-1.1\baselineskip}
\centering
  \includegraphics[width=0.95\linewidth]
  {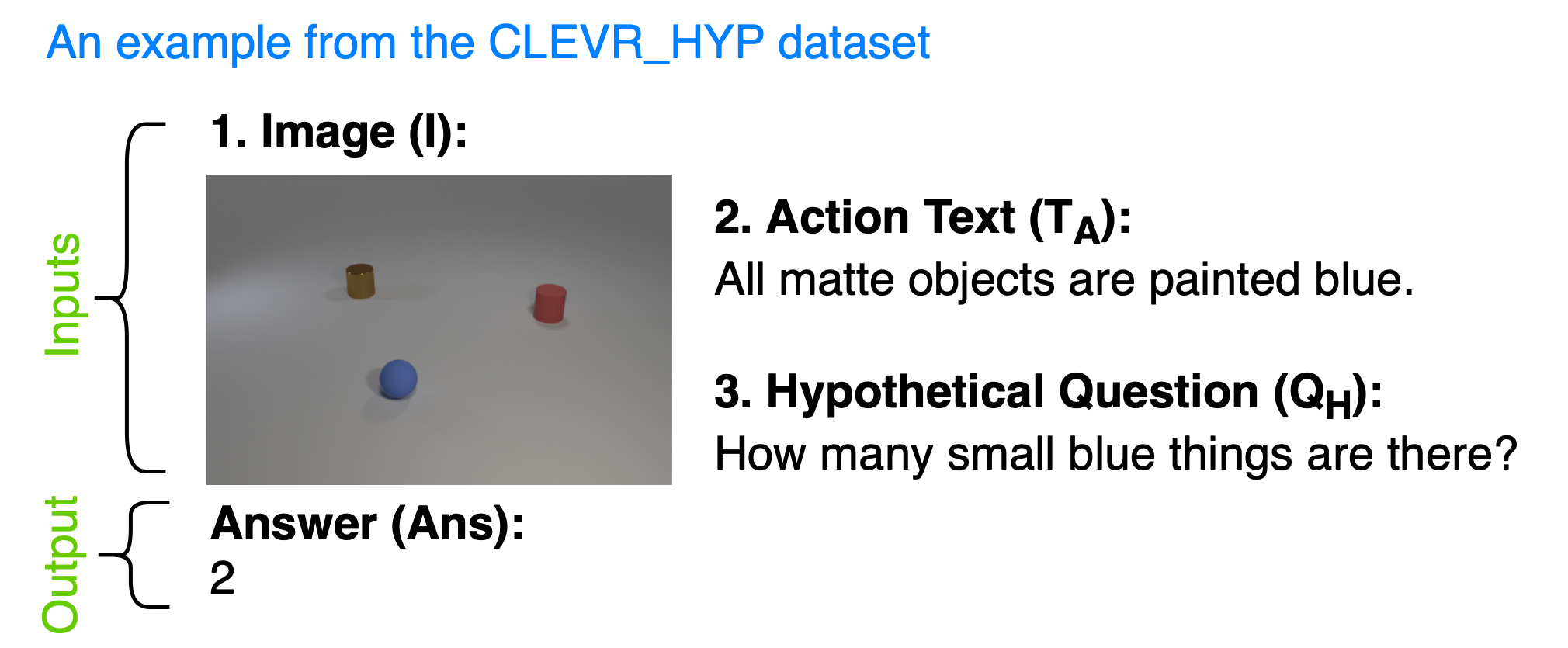}
  \captionof{figure}{An example from the CLEVR\_HYP dataset %task \cite{sampat2021clevr_hyp}: Answer a reasoning question (Q$_H$) about changes caused over the image (I) by performing a hypothetical action (T$_A$).
  }
  \label{fig:open_example}
  \end{figure}

The task aims at understanding changes caused over an image by performing an action described in natural language and then answering a reasoning question over the resulting scene. Figure \ref{fig:open_example} shows an example from the dataset;

\begin{itemize}[noitemsep,topsep=1pt,leftmargin=*]
\item \textit{Inputs:} 
\begin{enumerate}[noitemsep,topsep=0pt,leftmargin=*]
    \item Image (I)- Visual scene with rendered objects  
    \item Action Text (T$_A$)- Text describing action to be performed over I
    \item Question (Q$_H$)- Question to assess the system's ability to understand changes caused by T$_A$ on I
\end{enumerate}
\item \textit{Output:} Answer (A) for the given Q$_H$
\item \textit{Answer Vocabulary:} [0-9, yes, no, cylinder, sphere, cube, small, big, metal, rubber, red, green, gray, blue, brown, yellow, purple, cyan]
\item \textit{Evaluation:} 27-way Answer Classification / Accuracy (\%)
\end{itemize}
%\subsection{Example from the CLEVR\_HYP dataset} 
\subsection{Dataset Details and Partitions}
\label{sec:datapart}
The CLEVR\_HYP dataset assumes to have a closed set of object attributes, action types, and question reasoning types. 
\begin{itemize}[noitemsep,topsep=1pt,leftmargin=*]
\item Object attributes: 5 colors, 3 shapes, 2 sizes, 2 materials 
\item Action types: Add object, Remove object(s), Change attribute, Move object (in-plane and out-of-plane)
\item Reasoning types: Count objects, Compare objects, Existence of objects, Query attribute and Compare attribute
\end{itemize}
The dataset is divided into the following partitions;

\begin{itemize}[noitemsep,topsep=1pt,leftmargin=*]
    \item \textit{Train} (67.5k) /  \textit{Val} (13.5k) sets have $<$I, T$_A$, Q$_H$, A$>$ tuples along with the scene-graphs as a visual oracle and functional programs\footnote{Originally introduced in CLEVR \cite{johnson2017clevr}, ex. question `How many red metal things are there?' $\sim$ functional program `count(filter\_color(filter\_material(scene(),metal),red))'} as a textual oracle.  
    \item \textit{Test} sets consist of only $<$I, T$_A$, Q$_H$, A$>$ tuples, and \textit{no oracle annotations available}. There are three test sets,
    \begin{enumerate}[noitemsep,topsep=0pt,leftmargin=*]
    \item Ordinary test (13.5k) consists of examples with the same difficulty as train/val
    \item 2HopT$_A$ test (1.5k) consists of examples where two actions are performed ex. `Move a purple object on a red cube \textit{then} paint it cyan.'
    \item 2HopQ$_H$ test (1.5k) consists of examples where two reasoning types are combined ex. `How many objects are \textit{either} red \textit{or} cylinder?'
    \end{enumerate}
\end{itemize}

\subsection{\textbf{Baseline Models}}
\label{sec:bl}
%Following is a brief description of 
Following are two top-performing baselines reported in \citet{sampat2021clevr_hyp}, to which we will compare the results of our proposed approach in this paper. %More details about each baseline and appropriate hyperparameters are provided in Appendix \ref{sec:appendix1}.

\begin{itemize}[noitemsep,topsep=3pt,leftmargin=*]
    \item \textbf{(TIE) Text-conditioned Image Editing:} Text-adaptive encoder-decoder with residual gating \cite{vo2019composing} is used to generate new image conditioned on the action text. Then, new image along with the question is fed into LXMERT \cite{tan2019lxmert} (which is a pre-trained vision-language transformer), to generate an answer. %The model can be visualized in Figure \ref{fig:bl2}. %LXMERT is a pre-trained vision-language transformer which supports Visual Question Answering (VQA) downstream task.
    
    \item \textbf{(SGU) Scene-graph Update:} In this model, understanding changes caused by an action text is considered as a graph-editing problem. First, an image is converted into a scene-graph and action text is converted into a functional program (FP). \citet{sampat2021clevr_hyp} developed a module inspired by \citet{chen2020graph} 
    that can generate an updated scene graph provided the initial scene-graph and a functional program of action text. It is followed by a neural-symbolic VQA model \cite{yi2018neural} that can generate an answer to the question provided the updated scene-graph. %The model can be visualized in Figure \ref{fig:bl4}.
\end{itemize}

\section{Proposed Model: Action Representation Learner (ARL)}
\label{sec:arl}

\begin{figure*}
    \centering
  \includegraphics[width=0.95\linewidth]{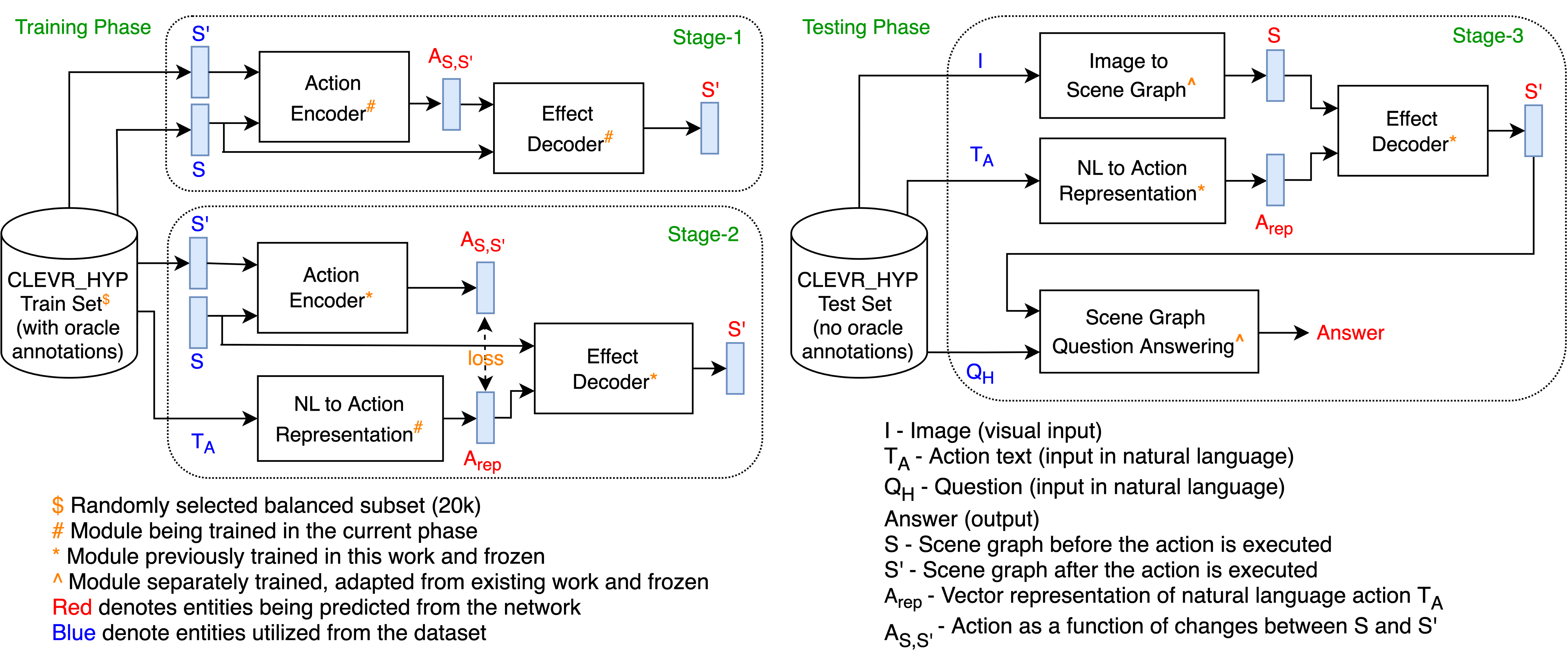} 
  \caption{Detailed visualization of our proposed 3-stage Action Representation Learner (ARL) model: (left) training phase (right) testing phase (bottom) terms and notations used. Best viewed in color. }
  \label{fig:reasoner}
\end{figure*}

In this section, we describe the architecture of our proposed model Action Representation Learner (ARL). Our hypothesis is that a model can learn better action representations by observing difference between a pair of states (before and after the action is performed) and then associate those visual differences with linguistic description of actions. We create a 3-stage model shown in Figure \ref{fig:reasoner}, which we believe would better capture the causal structure of this task.

\subsection{Stage-1}
This stage comprises an `Action Encoder' and an `Effect Decoder'. As described in previous section, the training set of CLEVR\_HYP provides oracle annotations for an initial scene-graph S (using which the image is rendered) and a scene-graph after executing the action text S$^{\prime}$. We take a random subset of 20k %\footnote{We experiment with different data sizes and discuss results in Section \ref{sec:abl}, but obtain the optimal results for 20k samples when action vector length is 125} 
 scene-graph pairs from the training set balanced by action types (add, remove, change, move) to train this encoder-decoder. 

We capture the difference between states S and S$^{\prime}$ i.e. A$_{S,S'}$ using the encoder. At the test time, we do not have the updated scene-graph S$^{\prime}$ available. To address this issue, the encoder is followed by a decoder, which can reconstruct S$^{\prime}$ provided S and the learned scene difference A$_{S,S'}$. We jointly train encoder-decoder with the following objective;
\begin{multline}
argmax_{\Theta _{ActionEncoder}\Theta _{EffectDecoder}}^{} \\  [ log P(\textcolor{red}{S^{\prime}}|\frac{}{}\textcolor{blue}{S},\frac{}{}ActionEncoder(\textcolor{blue}{S}, \textcolor{blue}{S^{\prime}})) ]
\end{multline}

\subsection{Stage-2}
\label{sec:parser}

%As explained in Stage-1, we do not have the S$^{\prime}$ at the test time and we cannot compute A$_{S,S'}$. However, if 
We assume changes in the scene as a function of action. However we can obtain A$_{rep}$ i.e. vector representation of natural language action, and it can be trained to approximate A$_{S,S'}$ (from Stage-1). Specifically, we freeze encoder-decoder trained in Stage-1 and learn `Natural language to Action Representation' module that maximizes the following log probability;

 \begin{multline}
argmax_{\Theta _{NL2ActionRep}}^{} \\  [ log P(\textcolor{red}{S^{\prime}}|\frac{}{}\textcolor{blue}{S},\frac{}{}NL2ActionRep(\textcolor{blue}{T_A})) ]
\end{multline}

Inside this module lies LSTM encoder, which precedes by an embedding layer and followed by dense layers. During training, in addition to finding the values for the weights of the LSTM and dense layers, the word embeddings for each word in the training set are computed. %This is achieved using nn.Embedding(vocabulary\_size, embedding\_size) layer defined in pytorch.
This way, a fixed length vector is generated for each word in the vocabulary depending on the position of the word in context and updated using back-propagation. The LSTM has a hidden layer of size 200. %Embedding layer is similar to a linear layer, which returns the index where one is located instead of returning the whole one-hot vector. 
%It takes an action text T$_A$ as a sequence of learned word embeddings, runs an LSTM over them, then projects from the final cell state to get the output A$_{rep}$. 

\subsection{Stage-3}
\label{sec:stage3}

Stage-3 combines modules trained in Stage-1 and Stage-2 with off-the-shelf `Image to Scene-graph' and `Scene-graph Question Answering' networks. Specifically, `Image to Scene-graph' is Mask R-CNN \cite{he2017mask} followed by a ResNet-34 \cite{he2016deep}, that classifies visual attributes- color, material, size, and shape and obtains 3D coordinates of each object in the scene. The `Scene-graph Question Answering' network is based on \cite{yi2018neural}, which has near-perfect accuracy on the scene-graph question answering task over CLEVR \cite{johnson2017clevr}. 

\section{Results and Analysis}

\subsection{Quantitative Results}

In this section, we discuss performance of our model quantitatively and qualitatively. We also discuss three ablations conducted for our model. 

\noindent\textbf{Evaluation Metric:} The classification task of CLEVR\_HYP has exactly one correct answer. Therefore, the exact match accuracy (\%) metric is used for evaluation. 

\begin{table}[ht!]
\centering
\begin{tabular}{llccc}
\hline
\multicolumn{5}{c}{\textbf{Test performance on CLEVR\_HYP(\%)}} \\ \hline
  &  & \textit{TIE} & \textit{SGU} & \textit{ARL} \\
\textit{Ordinary} &  & 64.7 & 70.5 & 76.4 \\
\textit{2HopA$_T$} &  & 55.6 & 64.4 & 69.2 \\
\textit{2HopQ$_H$} &  & 58.7 & 66.5 & 70.7 \\ \hline
\end{tabular}
\caption{Performance of two baselines (TIE, SGU) reported in \cite{sampat2021clevr_hyp} and our proposed model (ARL) on three test sets of CLEVR\_HYP}
\label{tab:quanr}
\end{table}

Our experimental results are summarized in Table \ref{tab:quanr}. %demonstrates the performance of our proposed model in comparison with the two best performing existing models TIE and SGU, described in Section \ref{sec:bl}. 
Our proposed approach (ARL) outperforms existing baselines by 5.9\%, 4.8\% and 4.2\% on \textit{Ordinary}, \textit{2HopT$_A$ Test} and \textit{2HopQ$_H$ Test} respectively. This demonstrates that our model not only achieves better overall accuracy but also has improved generalization capability when multiple actions have to be performed on the image or understand logical combinations of attributes while performing reasoning. 

\subsection{Qualitative results}
\begin{figure}[ht!]
 \setlength\belowcaptionskip{-1.1\baselineskip}
\centering
  \includegraphics[width=0.85\linewidth]{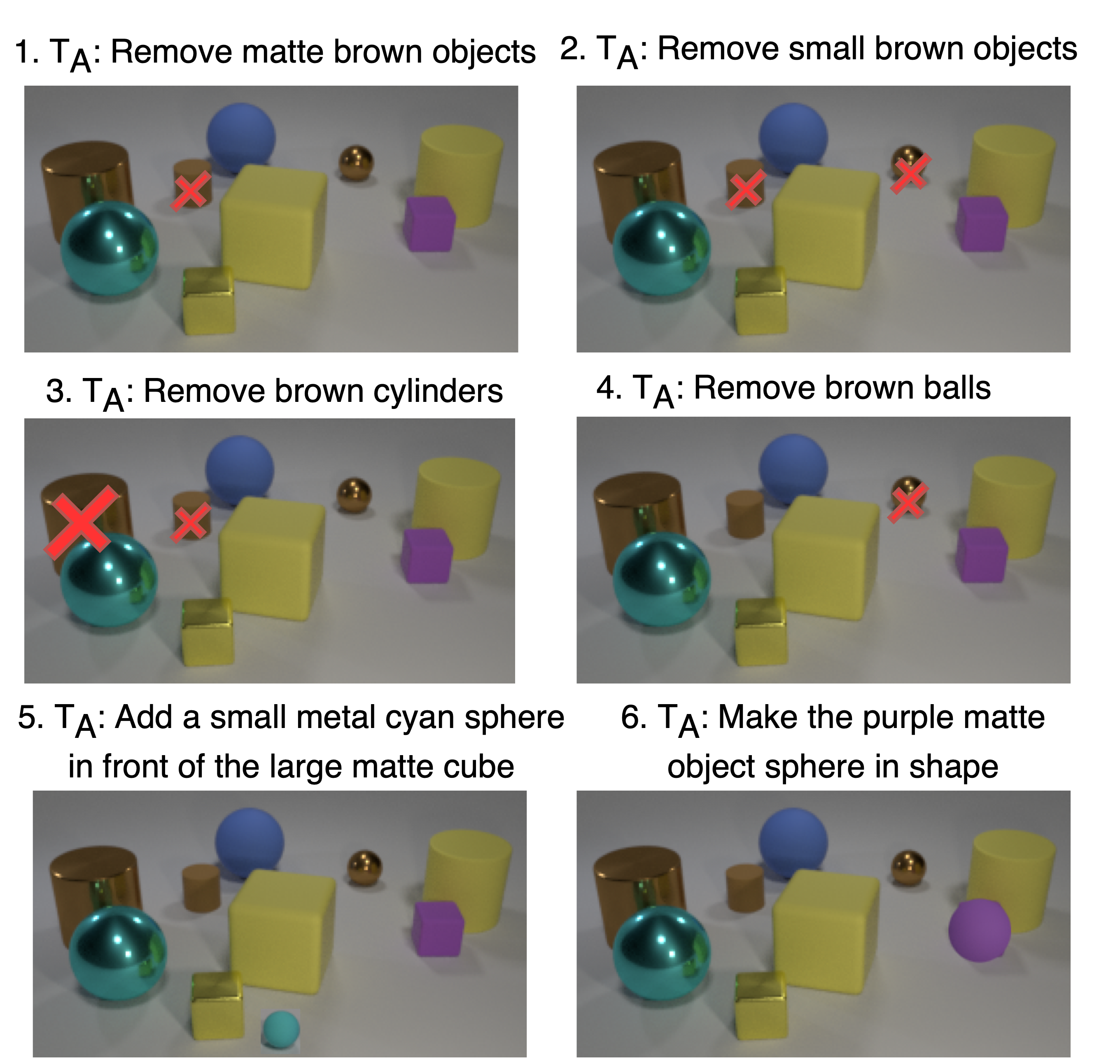} 
  \caption{Correct scene-graph predictions for the given input image and action text by our ARL model}
  \label{fig:goodex}
\end{figure}
In Figure \ref{fig:goodex}, we visually demonstrate scene-graphs predicted by our ARL model over a variety of action texts. From examples 1-2, we can observe that the model can correctly identify objects that match the object attributes (color, size, shape, material) provided in the action text. Examples 3-4 demonstrate that our system is consistent in predictions when we use synonyms of various words (e.g. sphere$\sim$ball, shiny$\sim$metallic) in the dataset. Finally, examples 5-6 show that our model does reasonably well on other actions (add and change).

\begin{figure}[ht!]
 \setlength\belowcaptionskip{-1.1\baselineskip}
\centering
  \includegraphics[width=0.7\linewidth]{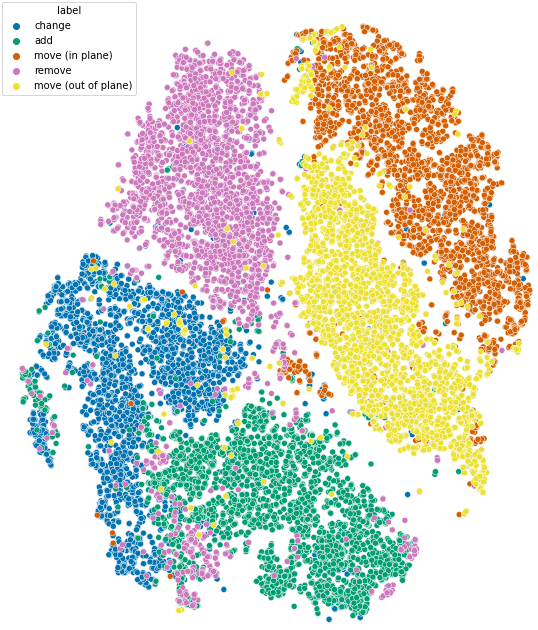}
  \caption{The t-SNE plot of learned action vectors}\label{fig:awesome_image1}
\end{figure}

We further generate a t-SNE plot of action vectors learned by our best proposed model, which is shown in Figure \ref{fig:awesome_image1}. At a first glance, we can say that the learned action representations formulate well-defined and separable clusters corresponding to each action type. Clusters for add, remove and change actions are closer and somewhat overlapping. %We observed that many samples of type `change' is  interpreted by the reasoner as `remove+add' action. For example, if a color of 'small blue metal sphere' is changed to 'red', the action reasoner interprets it as removal of the 'small blue metal sphere' followed by an addition of a 'small red metal sphere' on the same location. 

\subsection{Ablations}
\label{sec:abl}

\paragraph{Importance of Stage-1 training}
Cause-effect learning with respect to actions is a key focus in CLEVR\_HYP. In existing models, it is formulated as a updated scene-graph prediction task (i.e. given an initial scene and an action, determine what the resulting scene would look like after executing the action). In our opinion, Stage-1 plays a critical role in learning causal structure of the world. To demonstrate this, we set up two experiments; first, where training takes place in a sequential manner (Stage-1 followed by Stage-2), where trained encoder-decoders from Stage-1 are frozen and utilized in Stage-2. Second experiment, where there is no separate Stage-1 training and encoder-decoder in Stage-2 are randomly initialized. 

 \begin{figure}
  \setlength\belowcaptionskip{-0.9\baselineskip}
 \centering
 \includegraphics[width=0.8\linewidth]{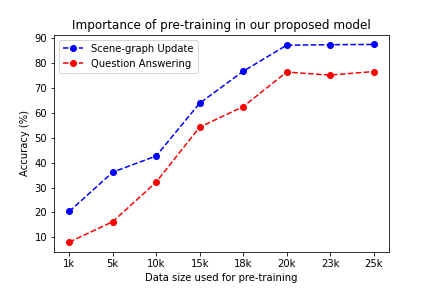}\\
  \includegraphics[width=0.8\linewidth]{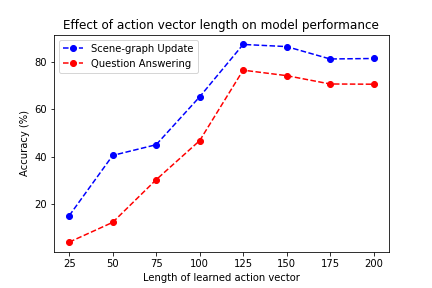}
  \caption{Performance of our model with varying (top) data size and (bottom) action vector lengths}
  \label{fig:graph1}
\end{figure}

\begin{table}[h!]
 \setlength\belowcaptionskip{-1.1\baselineskip}
 \centering
\begin{tabular}{@{}ccc@{}}
\toprule
\multicolumn{1}{l}{\textbf{Task}}                                              & \textbf{Experiment} & \textbf{Accuracy (\%)} \\ \midrule
\multirow{2}{*}{\begin{tabular}[c]{@{}c@{}}Scene-graph \\ Update\end{tabular}} & Stage(2 only)       & 56.3                   \\
                                                                               & Stage(1+2)          & 87.2                   \\ \midrule
\multirow{2}{*}{\begin{tabular}[c]{@{}c@{}}Question \\ Answering\end{tabular}} & Stage(2+3 only)     & 45.7                   \\
                                                                               & Stage(1+2+3)        & 76.4                   \\ \bottomrule
\end{tabular}
\caption{Performance of our model in the absence and presence of stage-1 over ordinary test set}
\label{tab:abl1}
\end{table}

The results are summarized in Table \ref{tab:abl1}. We can observe that inclusion of Stage-1 training improves the accuracy of scene-graph prediction by $\sim$30\% compared to the Stage(2 only) model. To evaluate question answering task of CLEVR\_HYP, both setups are followed by Stage-3 where the image to scene-graph generator and scene-graph question answering modules are combined to predict the answer. It is known that \cite{yi2018neural} has near-perfect performance on the scene-graph question answering task over CLEVR \cite{johnson2017clevr}. As a result, the gains achieved in the scene-graph task directly benefit the question answering performance without much of a loss. In other words, there are only  0.2\% instances where the scene prediction is correct but the final answer is incorrect. 

\paragraph{Performance with different data size used for training Stage-1}
In this ablation, the goal is to find out how many scene-graph pairs are required to effectively learn  effects of the actions. We experiment with different data sizes- from 2k to 25k samples. Figure \ref{fig:graph1} (top) shows the effect of training with diverse training data size on scene-graph update and downstream question answering task. The model learns better initially with more data samples, however performance saturates after 20k samples.

\paragraph{Performance with different lengths of learned action vector in Stage-1}
In this ablation, the goal is to find out optimal length of action vectors that can reasonably simulate the effects of the actions. We experiment with different lengths of learned action vector- from 25 to 200 in increment of 25. Figure \ref{fig:graph1} (bottom) shows the effect of training with diverse action vector lengths on scene-graph update and downstream question answering task. The model learns better initially when the vector length is increased, however performance reaches at peak for the action vector length of 125.

\section{Conclusion}

In the vision and language domain, several tasks are proposed that require an understanding of the causal structure of the world. In this work, we propose an effective way of learning action representations and implement a 3-stage model for the what-if vision-language reasoning task CLEVR\_HYP. We provide insights on the learned action representations and validate the effectiveness of our proposed method through ablations. Finally, we demonstrate that our proposed method outperforms existing baselines while being data-efficient and showing some degree of generalization capability. By extending our approach to a larger set of actions, we aim to develop AI agents which are equipped with action-effect reasoning capability and can better collaborate with humans in the physical world. 

\bibliography{aaai23}

\end{document}